\documentclass[10pt,twocolumn,letterpaper]{article}

\usepackage{cvpr}              

\usepackage{graphicx}
\usepackage{amsmath}
\usepackage{amssymb}
\usepackage{booktabs}
\usepackage{bbding}
\usepackage[accsupp]{axessibility}

\usepackage[pagebackref,breaklinks,colorlinks]{hyperref}

\usepackage[capitalize]{cleveref}
\crefname{section}{Sec.}{Secs.}
\Crefname{section}{Section}{Sections}
\Crefname{table}{Table}{Tables}
\crefname{table}{Tab.}{Tabs.}

\def\cvprPaperID{6289} 
\def\confName{CVPR}
\def\confYear{2022}

\title{Feature Erasing and Diffusion Network for Occluded Person Re-Identification}

\begin{document}

\author{Zhikang Wang\textsuperscript{1,2}\thanks{Zhikang Wang did this work as an intern in SenseTime Research.} , 
Feng Zhu\textsuperscript{2}\thanks{Corresponding author.} , Shixiang Tang\textsuperscript{3}, Rui Zhao\textsuperscript{2,4}, Lihuo He\textsuperscript{5} \thanks{This research was supported partially by the National Natural Science Foundation of China (Grant Nos. 61876146). }, Jiangning Song\textsuperscript{1} \\
\textsuperscript{1}Monash University, \textsuperscript{2}SenseTime Research, \textsuperscript{3}The University of Sydney, \\ 
\textsuperscript{4}Qing Yuan Research Institute, Shanghai Jiao Tong University, \textsuperscript{5}Xidian University \\
{\tt\small zkwang00@gmail.com, zhufeng@sensetime.com, stan3906@uni.sydney.edu.au, zhaorui@sensetime.com,}\\ {\tt\small lihuo.he@gmail.com, jiangning.song@monash.edu}
}

\maketitle


\begin{abstract}
Occluded person re-identification (ReID) aims at matching occluded person images to holistic ones across different camera views. Target Pedestrians (TP) are often disturbed by Non-Pedestrian Occlusions (NPO) and Non-Target Pedestrians (NTP). Previous methods mainly focus on increasing the model's robustness against NPO while ignoring feature contamination from NTP. In this paper, we propose a novel Feature Erasing and Diffusion Network (FED) to simultaneously handle challenges from NPO and NTP. Specifically, aided by the NPO augmentation strategy that simulates NPO on holistic pedestrian images and generates precise occlusion masks, NPO features are explicitly eliminated by our proposed Occlusion Erasing Module (OEM). Subsequently, we diffuse the pedestrian representations with other memorized features to synthesize the NTP characteristics in the feature space through the novel Feature Diffusion Module (FDM). With the guidance of the occlusion scores from OEM, the feature diffusion process is conducted on visible body parts, thereby improving the quality of the synthesized NTP characteristics. We can greatly improve the model's perception ability towards TP and alleviate the influence of NPO and NTP by jointly optimizing OEM and FDM. Furthermore, the proposed FDM works as an auxiliary module for training and will not be engaged in the inference phase, thus with high flexibility. Experiments on occluded and holistic person ReID benchmarks demonstrate the superiority of FED over state-of-the-art methods. 
\end{abstract}

\begin{figure}[t]
  \centering
   \includegraphics[width=1\linewidth]{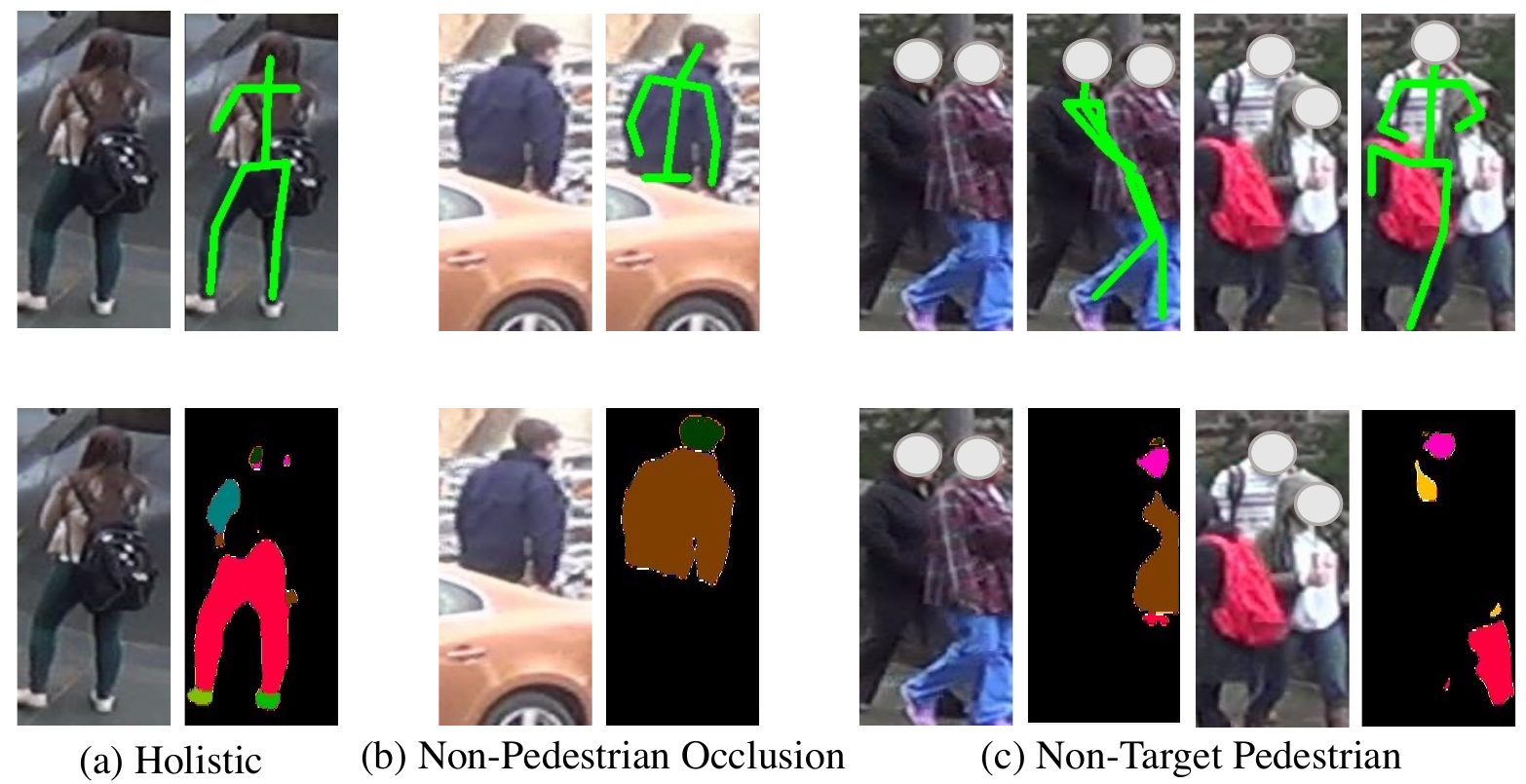}
   \caption{Illustration of pose estimation and human parsing on pedestrian images. Both models perform well on holistic and object occluded pedestrians but fail on multi-pedestrian images. Meanwhile, human parsing models have difficulty in identifying personal belongings, \emph{e.g.}, backpacks, and umbrellas. 
   }
   \label{Fig1}
\end{figure}

\section{Introduction}
Person Re-Identification (ReID) aims at retrieving the same pedestrians captured by different cameras with different viewpoints, lighting conditions, and locations. With the rapid development of deep learning area and publication of large-scale image and video ReID datasets, ReID methods based on deep neural networks have achieved remarkable performance \cite{hermans2017defense,luo2019bag,wang2021robust,wang2019multi}. Most of these approaches assume that a holistic body of each pedestrian is available for feature extraction. However, in real-world scenarios, \emph{e.g.}, railway stations, schools, hospitals, and shopping malls, pedestrians are inevitably disturbed by non-pedestrian occlusions (NPO) and non-target pedestrians (NTP). Therefore, designing a powerful network for the occluded person ReID is essential. 

Methods assisted by human key points \cite{gao2020pose,miao2019pose} and human parsing information \cite{huang2020human} dominate the state-of-the-art performance of the occluded ReID task. Generally, an auxiliary model extracts the body information first, and then the extracted information will assist the training of models. The strategy can greatly avoid mistakenly treating NPO as human parts. However, such methods have many caveats. Firstly, due to the domain gap between the training and testing data, the performance of the auxiliary models can not be consistent. In Fig.\ref{Fig1}, we adopt official pose estimation model \cite{sun2019deep} and retrained human parsing model \cite{zhao2017pyramid} to extract body information. It is clear that both models perform well on holistic and object occluded pedestrian images but fail on multi-pedestrian ones, which means that noise from NTP will contaminate the final representations. Compared with object occlusion, the characteristics of NTP will result in a higher mismatching probability because of the semantic guidance. Secondly, the human parsing model can not recognize some person belonging, \emph{e.g.}, backpacks, umbrellas, which may lead to the deficiency of valuable information. At last, the enormous computation brought by the auxiliary models makes it unacceptable for real-time video surveillance.

To tackle the challenges above, we propose the feature erasing and diffusion network (FED) to simulate NPO on images and NTP in the feature space for increasing the model's perception ability towards TP. Specifically, we aim at the NPO feature erasing by proposing the NPO augmentation strategy along with an occlusion erasing module (OEM). 
The augmentation strategy will generate object occluded data of pedestrians by pasting cropped patches with a specific strategy. At the same time, by analyzing the pixel-level value differences, we can get precise part labels, indicating whether object occlusion or not. We refer to the part labels as occlusion masks. Sequentially, the occlusion masks will guide the OEM to analyze the semantic information and generate the final occlusion scores for part features. For alleviating the distractions from NTP, a straightforward way is pasting other pedestrians onto the image for data augmentation. However, pedestrian images with diversified background information can destroy the globality of the original images by simple pasting. Besides, the resize operation needs designing carefully for maintaining aspect ratio. Therefore, image-level augmentation for NTP is challenging and complex. Here, we propose a learnable structure named feature diffusion module (FDM), which will simulate multi-pedestrian images by diffusing characteristics of NTP to the original features. With the guidance of occlusion scores from OEM, the feature diffusion operation will be conducted only on body parts,  guaranteeing the simulated features are more realistic. By optimizing the model through diffused features, we can indirectly improve the model's perception ability towards TP and robustness towards NTP.

In summary, we propose the feature erasing and diffusion network (FED) to tackle the distractions from NPO and NTP for occluded person ReID. FED consists of three innovative components: NPO augmentation strategy, occlusion erasing module (OEM), and feature diffusion module (FDM). These components enable the network to precisely perceive the TP regardless of the NPO and NTP. At the same time, extensive experiments on both occluded datasets (Occluded-DukeMTMC \cite{miao2019pose}, Partial-REID \cite{yang2019attention}, and Occluded-REID \cite{zhuo2018occluded}) and holistic datasets (Market-1501 \cite{zheng2015scalable} and DukeMTMC-reID \cite{zheng2017unlabeled}) demonstrate the effectiveness of our proposed method. Especially on the Occluded-DukeMTMC and Occluded-REID dataset, our Rank-1 and mAP accuracy surpass other state-of-the-art methods by a large margin.

\begin{figure*}[t]
  \centering
   \includegraphics[width=0.9\linewidth]{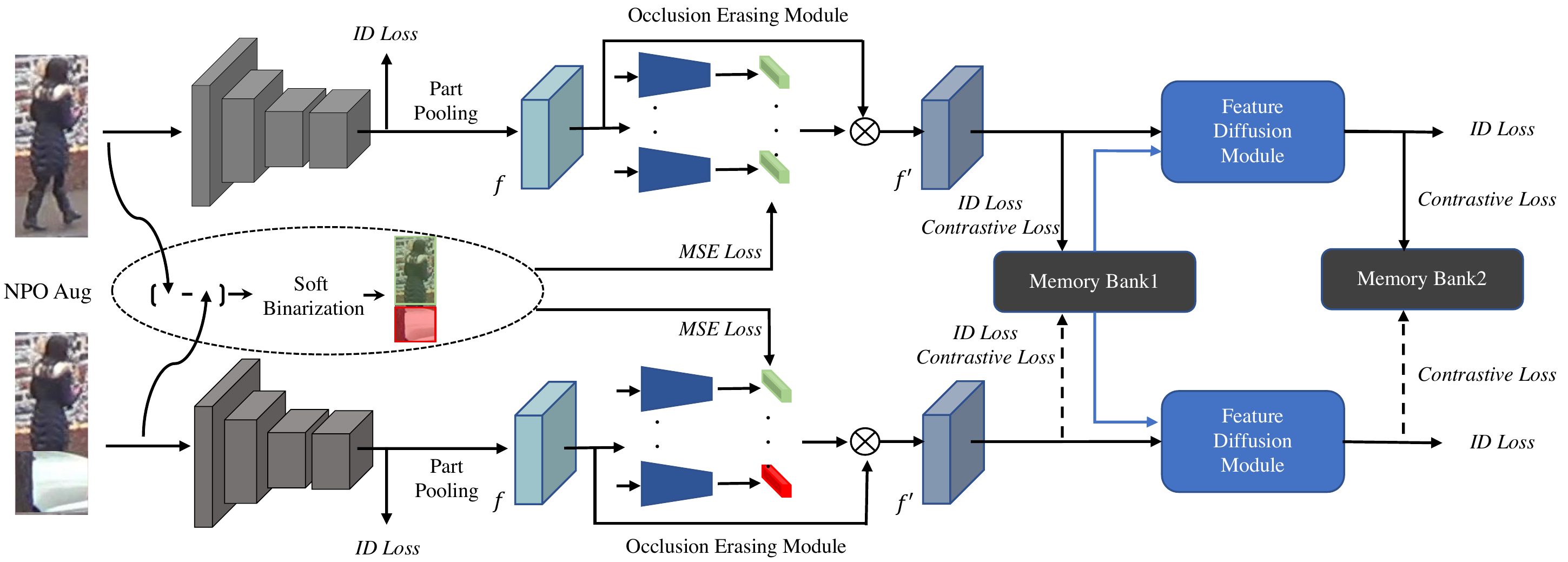}
   \caption{Overview of the proposed feature erasing and diffusion network for occluded person re-Identification. The two branches share the same parameters and the network consists of the feature extractor, occlusion erasing module (OEM), and feature diffusion module (FDM). The `NPO Aug' indicates the NPO augmentation strategy. The solid lines connected to the Memory Banks indicate that the features participate in the memory update and loss calculation. The dashed lines indicate only loss calculation. The FDM is an auxiliary module for simulating NTP on feature level and will not be engaged in the inference phase. 
   }
   \label{Fig2}
\end{figure*}

\section{Related Works}
In this section, we briefly overview the existing methods of holistic person ReID and occluded person ReID. 

\subsection{Holistic Person Re-Identification}
Person re-identification (ReID) aims to retrieve a person of interest in other camera views and great progress has been made in recent years. Existing ReID methods can be summarized into three categories, including hand-crafted descriptor methods\cite{yang2014salient,ma2014covariance}, metric learning methods\cite{zhong2017re,chen2017beyond}, and deep learning methods\cite{wang2020robust,sun2018beyond,xu2018attention}. Due to the publishing of large-scale datasets and the development of Graphics Processing Unit (GPU), deep learning based methods have become dominant in the person re-identification area nowadays. 
Recent works utilizing part-based features have achieved state-of-the-art performance for the holistic person ReID. Zhang \emph{et al.} \cite{zhang2017alignedreid} perform an automatic part feature alignment through the shortest path loss during the learning, without requiring extra supervision or explicit pose information. Sun \emph{et al.} \cite{sun2018beyond} propose a general part-level feature learning method, which can accommodate various part partitioning strategies. 
The attention mechanism has also been adopted to ensure the model focus on human areas, which extracts more effective features \cite{li2018harmonious,tay2019aanet,yang2019attention}. 
However, these methods fail to retrieve persons with high accuracy when occlusions happen. The shortcoming limits the utility of the methods, especially in the common crowd scenes.

\subsection{Occluded Person Re-Identification}
The study of the occluded person ReID is proposed by Zhou \emph{et al.} \cite{zhuo2018occluded}. The training set and gallery set are constructed by holistic pedestrian images, and the query set is constructed by occluded pedestrian images. Recent study methods in this topic can be divided into two categories: assisted by pose estimation \cite{he2019foreground,he2020guided} and human parsing \cite{huang2020human,yu2021neighbourhood}.
Gao \emph{et al.} propose a Pose-guided Visible Part Matching (PVPM) method that jointly learns the discriminative features with pose-guided attention and self-mines the part visibility in an end-to-end framework. He \emph{et al.} \cite{he2020guided} introduce a novel method named Pose-Guided Feature Alignment (PGFA), exploiting pose landmarks to disentangle the useful information from the occlusion noise.
Huang \emph{et al.} propose a model named HPNet to extract part-level features and predict the visibility of each part, based on human parsing. By extracting features from semantic part regions and performing comparisons with consideration of visibility, the method not only reduces background noise but also achieves body alignment.

Different from the above methods, our approach does not rely on extra models and can be trained in an end-to-end fashion. We simulate NPO and NTP on both image and feature levels and thus greatly improve the model robustness.

\section{Feature Erasing and Diffusion Network}
In this section, we introduce the proposed feature erasing and diffusion network (FED) in detail. The overall architecture of the network is illustrated in Fig.\ref{Fig2}. 
It begins with the NPO augmentation strategy that produces image pairs and occlusion masks. Following \cite{he2021transreid}, we simply adopt the Vision Transformer (ViT)\cite{dosovitskiy2020image} as the feature extractor. Position embeddings and a classification [cls] token are prepended to the input image. 
The output feature for each image is $f \in \mathbb{R}^{(n+1) \times c}$, where $n+1$ indicates the images tokens and one [cls] token, and $c$ is the channel dimension. Under our settings, $n$ and $c$ are 128 and 768, respectively. 
Next, we conduct the part pooling operation on image tokens and obtain $N$ local features, which will be fed into the occlusion erasing module (OEM). Here, we set $N$ as 4 in accordance with NPO augmentation strategy. Two memory banks will be initialized at the beginning and updated with training processing. The auxiliary feature diffusion module (FDM) takes the image features and the first memory bank as input for multi-pedestrian simulation. Details of each module will be presented in the following section.

\subsection{NPO Feature Erasing}
Typically, NPO feature erasing needs auxiliary information for guidance. In this section, we propose the NPO augmentation strategy and occlusion erasing module to explicitly learn NPO-robust features. 

\textbf{NPO Augmentation Strategy.}
Occlusion augmentation strategies are effective in occluded ReID. Typically, there are two categories: (1) Zhong \emph{et al.} \cite{zhong2020random} randomly select a rectangle region in an image and erase its pixels with random values; (2) Chen \emph{et al.} \cite{chen2021occlude} paste the selected objects or backgrounds onto images. The first method helps to reduce the risk of over-fitting and makes the model robust to occlusion. However, when facing the diversified occlusions, the trained model fails to identify them due to weak generalization. The second method implicitly learns NPO-robust features by simulating the occlusion scenes. However, it fails to fully utilize the potential information, \emph{e.g.}, precise occlusion region, brought by the augmentation.

Inspired by the methods above, we propose the NPO augmentation strategy. The strategy consists of occlusion augmentation and mask generation, which will generate augmented images for occlusion simulation and occlusion masks for further semantic analysis, respectively.

Empirically, occlusions happen at four locations (top, bottom, left, right) with a quarter to half areas. Our augmentation strategy is similar to Chen \emph{et al.} \cite{chen2021occlude}, but with particular modifications. For occlusion augmentation, one important step is the patch set collection. To avoid extra body parts included in the patch set, we manually crop the backgrounds and occlusion objects from the chosen images in the training set and refer to these patches as the occlusion set. 
We formally describe the occlusion augmentation process as follows. \emph{Firstly}, given an input image, we do common augmentations, \emph{e.g.,} resize, padding, and random crop, on it and get $x \in \mathbb{R}^{3 \times h \times w}$, where $h$ and $w$ represent the height and width, respectively.
\emph{Secondly}, we select a patch $p \in \mathbb{R}^{3 \times p_h \times p_w}$ from the occlusion set, where $p_h$ and $p_w$ are the height and width. Rather than randomly paste the patch onto $x$, we believe that only reasonable occlusions for pedestrians can generate valuable data for training. 
Therefore, we calculate the aspect ratio of the patch: $\alpha = p_h / p_w$. When $\alpha$ is larger than 3, it implies the patch is more like a vertical occlusion, otherwise horizontal occlusion. 
Common augmentations, \emph{e.g.}, random crop, and colorjitter, are also applied on the patch for increasing its varieties. We resize the patch according to the occlusion type (horizontal or vertical) to $\mathbb{R}^{(H/4 \sim H/2, W)} $ and $\mathbb{R}^{(H, W/4 \sim W/2)}$, respectively. \emph{Thirdly}, we randomly select one corner of $x$ as the starting point and past the augmented patch on it. The augmented image is named $x'$.

Mask generation is a fine-to-coarse process. \emph{Firstly}, we get the pixel differences by subtraction and absolute function $d = |x-x'|$. 
Considering the subsequent part-based occlusion erasing module, each position of the occlusion mask should correspond to specific body parts. 
However, there are mis-alignments of semantics (body parts) between different images, fine-grained occlusion masks will have many false labels. Therefore, we roughly split the image into 4 stripes horizontally and aim at labeling them.
As said before, there are vertical and horizontal occlusions in real-world scenarios. 
Vertical occlusion only damages parts of the symmetric characteristics. Usually, ReID models can easily distinguish between pedestrians and vertical occlusions and get discriminative representations without referring to further information. Therefore, the vertical occlusion is ignored while mask generation and stripes are regarded as a human part (valued 1).
For the horizontal occlusion augmentation, we conduct the soft binarization operation. We take stripes covered more than three-quarters as occlusions (value 0), otherwise as human parts (value 1). In this way, we get the precise occlusion masks for the image pair.

\textbf{Occlusion Erasing Module.}
Although the augmentation strategy is employed while training, the NPO may still contaminate representations. To further eliminate the influence of NPO, we propose the occlusion erasing module (OEM) for part feature erasing. 
As shown in Fig.\ref{Fig2}, the module is constructed by 4 sub-modules corresponding to each body part. For each sub-module, it is constructed by two fully connected (FC) layers, one layer normalization \cite{ba2016layer}, and one $Sigmoid$ function. The layer normalization is placed between the FC layers, and the $Sigmoid$ function is located at the end. The first FC layer compresses the channel dimension to the quarter of the original one, aiming to wipe off the characteristic information and reserve the semantic ones. The final $Sigmoid$ function will output the regressed occlusion scores $s_i$ for each part feature. We refer to the multiplication between the occlusion scores and part features as $f'$. Functionally the progress can be represented by
\begin{equation}
    f_i'=Sigmoid(W_{rg}LN(W_{cp}f_i)) \cdot f_i,
\end{equation}
where $W_{cp} \in \mathbb{R}^{c/4 \times c}$, $W_{rg} \in \mathbb{R}^{1 \times c/4}$, $LN$ is the layer normalization and $i$ indicates $i$\_th part feature.

Here, the occlusion masks from the NPO augmentation strategy are adopted to supervise the training of OEM. We calculate the \emph{Mean Square Error (MSE) Loss} between occlusion masks and occlusion scores, and the function can be expressed as
\begin{equation}
    \mathcal{L}_{MSE}=\frac{1}{N}\sum_{i=1}^{4}(s_i, mask_i).
\end{equation}

\begin{figure}[t]
  \centering
   \includegraphics[width=0.9\linewidth]{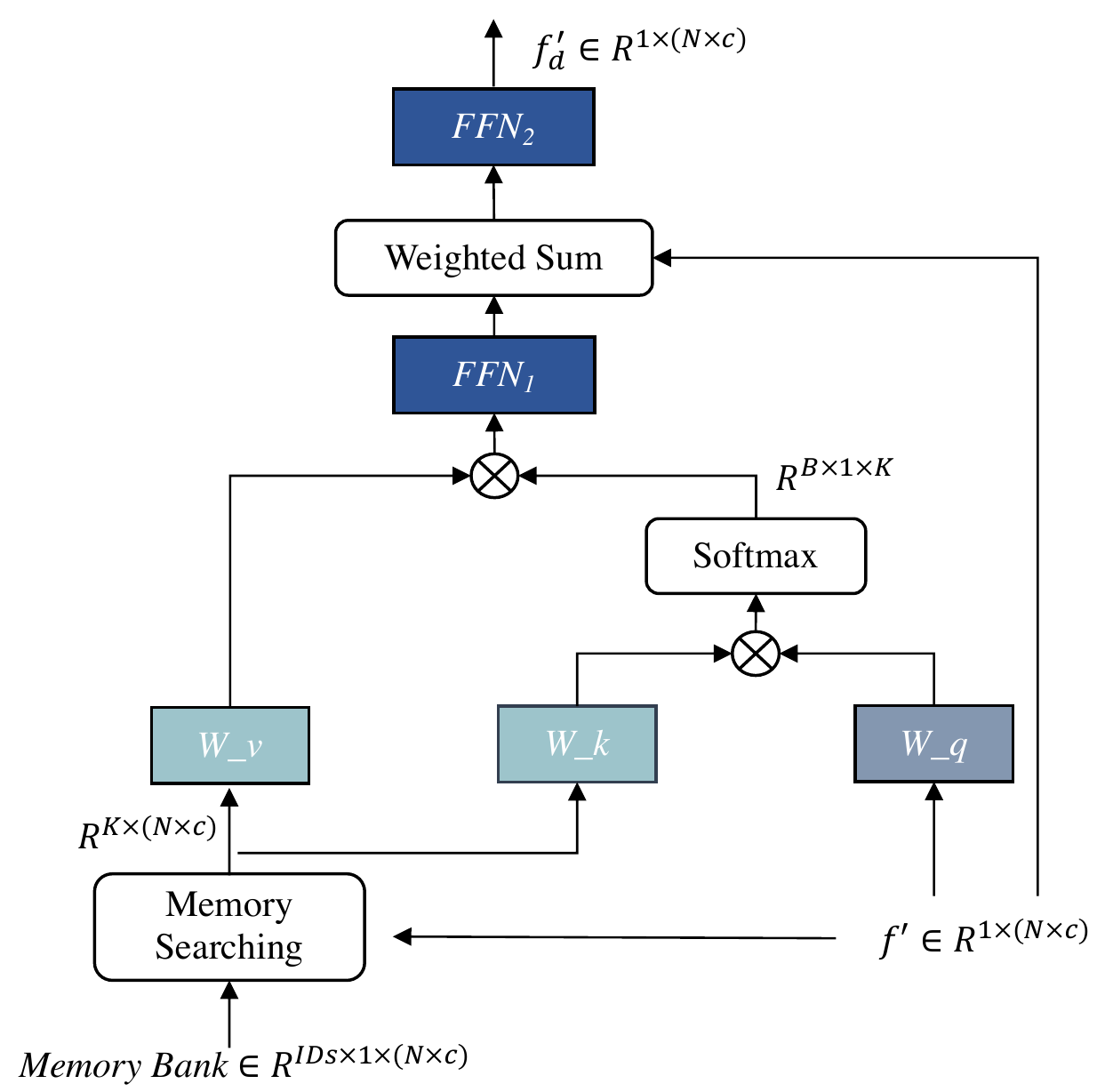}
   \caption{Illustration of the feature diffusion module. The module diffuses characteristics of memory bank $\mathcal{M}$ to the features $f'$ for simulating NTP on feature level. 
   }
   \label{Fig3}
\end{figure}

\subsection{Feature Diffusion Module}
Previous works have not focused on the challenges of NPO. Apart from destroying the feature integrity of the TP, NTP also contaminates representations with realistic semantic noise. To solve this issue, we propose a learnable structure named feature diffusion module (FDM) to simulate multi-pedestrian images in the feature space. By optimizing the diffused features, we aim at indirectly enhancing the model's perception ability towards TP and robustness towards NTP. As shown in Fig.\ref{Fig3}, apart from the image features, an extra memory bank$\mathcal{M}$, which is a collection of characteristics, is taken as the input. In the following session, we will introduce $\mathcal{M}$ and FDM, respectively. 

\textbf{Memory Bank.}
The generation of $\mathcal{M}$ includes memory initialization and memory update. We follow the same strategy as \cite{ge2020self}. The memory is initialized with the ID centers in the training set. We get the extracted features by performing forward computation, and average features with identical identities to get ID centers. 
Note that the memory initialization is only operated at the beginning of the algorithm and memory update is processed at each iteration in each mini-batch during training. The $k$-th center $c_k$ is updated by the mean of the encoded features belonging to identity $k$ in the mini-batch as:
\begin{equation}
    c_k = m c_k + (1 - m) \frac{1}{|B_k|}\sum_{f_i' \in B_k}f_i',
\end{equation}
where $B_k$ denotes the feature set belonging to identity $k$ in the mini-batch, $m$ is the momentum coefficient for updating, $f'$ is the flattened features after OEM. Apart from acting as the characteristic set, the memory bank $\mathcal{M}$ is also adopted for calculating the \emph{Contrastive Loss} which will be introduced in the following section. 
We set $m$ as 0.2 in our experiments. 

\textbf{Feature Diffusion Module.}
Essentially, FDM is a modified cross attention module based on the standard architecture of the transformer\cite{vaswani2017attention}. 
Given the feature vector, \emph{queries} $Q$ arise from the $f'$, and \emph{keys} $K$ and \emph{values} $V$ arise from the memory bank $\mathcal{M}$. 
The input feature is $f' \in \mathbb{R}^{1 \times (N \times c)}$, where $N$ corresponds to the previous part pooling operation and is 4.
Firstly, we conduct Memory Searching Operation between $f'$ and $\mathcal{M}$. It finds $K$ nearest centers $\mathcal{M}^K \in \mathbb{R}^{K \times (N \times c)}$ with different identities from the input image. Cosine distance is adopted for measurement. Here, we discard the center with an identical identity for avoiding polarization of the attention matrix which is calculated through cross-product. 
Formally, 
\begin{equation}
    Q = f' W^1, K_i = \mathcal{M}^K_i W^2, V_i = \mathcal{M}^K_i W^3,
\end{equation}
where $i\in1,2,...,K$, and $W_1 \in \mathbb{R}^{d \times d'}$, $W_2 \in \mathbb{R}^{d \times d'}$, $W_3 \in \mathbb{R}^{d \times d'}$ are linear projections. Then we calculate the attention matrix and corresponding part features. Formally, 
\begin{equation}
   m_{i}=\frac{exp(\beta_{i})}{\sum_{j=1}^{K}exp(\beta_{j})}, \quad \beta_i=\frac{QK_i}{\sqrt{d_k}},
\end{equation}
where $\sqrt{d_k}$ is a scaling factor. Each element of the attention matrix indirectly indicates the connections between $Q$ and $K_i$, and the cross-product operation between $V$ and the attention matrix will generate the diffused features. 
The aggregation process can be defined as:
\begin{equation}
    f_{d} = Att(Q, K, V) = \sum_{i=1}^{K}m_i V_i,
\end{equation}
The multi-head attention operation is of great significance in this module. Since $\mathcal{M}^K$ has many similar patterns with the input image and these patterns are distributed randomly in $K$ feature centers. 
The multi-head operation will split each center into multi parts and generate attention weights for each part individually, thus ensuring more patterns similar to TP and sufficient unique patterns of NTP can be aggregated. 
In this way, we can simulate the multi-pedestrian images on feature level. 
After the cross attention operation, we utilize the post-layer normalization feed forward network ($FFN_1$) \cite{xiong2020layer} to conduct non-linear transformation. $FFN_1(\cdot)$ is a simple neural network with two fully connected layers and one activation function. The residual connection before the layer normalization is applied. 
Next, the occlusion scores generated by OEM are adopted for weighted summation between the transformed features and $f'$. 
This ensures the characteristics of NTP are only added on human parts rather than pre-recognized object occlusion parts, improving the realness and quality of the diffused features. Besides, the weighted residual operation can stabilize the training process. 
Then, we utilize another $FFN_2$ \cite{xiong2020layer} for generating the final diffused representation of each image. 
Formally,
\begin{equation}
    f_{d}' = FFN_2(mask \cdot FFN_1(f_{d}) + f'),
\end{equation}
where $FFN_2$ has the same structure as $FFN_1$.

Since the FDM is just an auxiliary module for simulation during training, it will be removed in the inference phase. This makes our model more concise and flexible.




\subsection{Loss Functions}
There are three varieties of loss functions in our method, including \emph{Mean Square Error (MSE) Loss}, \emph{Cross Entropy Loss}, and \emph{Contrastive Loss}. We refer to \emph{Cross Entropy Loss} as \emph{ID Loss} in this paper. As shown in Fig.\ref{Fig2}, we calculate the \emph{ID Loss} on the output features of the classification [cls] token, flattened features after the OEM, and features after the FDM. Therefore, there are three additional fully connected layers on the top of the features to calculate the ID probabilities. 
Functionally, \emph{ID} Loss can be presented as:
\begin{equation}
    \mathcal{L}_{ID}=-y_ilog(\frac{exp(W_if_i)}{\sum_{j=1}^{ID_s}exp(W_jf_j)}),
\end{equation}
where $W$ is a linear projection matrix, $y_i$ is the corresponding label and $IDs$ is the total number of identities. 
As for the \emph{Contrative Loss}, the key components are the negative and positive samples. There are two memory banks in our algorithm, the first is generated after the OEM and the second is generated after FDM. The initialization and update strategies have been introduced in \emph{Sec 3.2}. 
Functionally, the \emph{Contrative Loss} is:
\begin{equation}
    \mathcal{L}_{C}=-log\frac{exp(<f,c_i>/\tau )}{\sum_{j}^{IDs}exp(<f,c_j>/\tau)},
\end{equation}
where $\tau$ is a predefined temperature parameter and $c_i$ represents the feature center with an identical identity. 
Although the training strategy is a parallel architecture, the lower branch does not involve in the memory initialization and update due to the characteristic deficiency caused by the NPO augmentation. In Fig.\ref{Fig2}, we utilize the solid lines to represent jointly memory update and loss calculation and dashed lines to represent loss calculation only. 

Therefore, the final loss function can be expressed:
\begin{equation}
    \mathcal{L}_{Final}=\frac{1}{2}\sum_{i=1}^{2}\mathcal{L}_{MSE}^i + \frac{1}{2}\sum_{i=1}^{6}\mathcal{L}_{ID}^i + \frac{1}{2}\sum_{i=1}^{4}\mathcal{L}_{C}^i.
\end{equation}

\section{Experiments}
\subsection{Datasets and Evaluation Setting}
\textbf{Occluded-DukeMTMC} \cite{miao2019pose} consists of 15,618 training images of 702 persons, 2,210 query images of 519 persons, and 17,661 gallery images of 1,110 persons. It is the most challenging occluded person ReID datasets due to the diverse scenes and distractions.

\textbf{Occluded-REID} \cite{zhuo2018occluded}  is an occluded person ReID dataset captured by mobile cameras. It consists of 2,000 images belonging to 200 identities. Each identity has five full-body person images and five occluded person images with different viewpoints and different types of severe occlusions.

\textbf{Partial-REID} \cite{zheng2015partial} is a specially designed ReID dataset that consists of occluded, partial, and holistic pedestrian images. It involves 600 images of 60 persons. We take the occluded query set and holistic galley set for experiments. 

\textbf{Market-1501} \cite{zheng2015scalable} is a famous holistic person ReID dataset. It contains 12,936 training images of 751 persons, 19,732 query images and 3,368 gallery images of 750 persons captured from 6 cameras. Few images in this dataset are occluded. 

\textbf{DukeMTMC-reID} \cite{zheng2017unlabeled} consists of 16,522 training images of 702 persons, 2,228 queries of 702 persons, and 17,661 gallery images of 702 persons. The images are captured by 8 different cameras, making it more challenging. As it contains more holistic images than occluded ones, this dataset can be treated as a holistic ReID dataset.

\textbf{Evaluation Protocol.}
To guarantee a fair comparison with existing person ReID methods, all methods are evaluated under the Cumulative Matching Characteristic (CMC) and mean Average Precision (mAP). All experiments are performed in the single query setting.

\subsection{Implementation Details}
Unless otherwise specified, all images are resized to $256 \times 128$. We train our network in an end-to-end fashion through the SGD optimizer with a momentum of 0.9 and weight decay of 1e-4. We initialize the learning rate as 0.008 with cosine learning rate decay. For each input branch, the batch size is 64, which contains 16 identities and 4 samples per identity. We conduct all experiments on two RTX 1080Ti GPUs. We set the temperature $\tau$ in \emph{Contrastive Loss} as 0.05 and the number of heads in the FDM as 8. 
For the occlusion set of NPO augmentation, we crop 30 patches from the training data of Occluded-DukeMTMC and MSMT17 \cite{wei2018person} as occlusion set 1 ($OS_1$) and occlusion set 2 ($OS_2$), respectively. If not specified, we only adopt $OS_1$ for NPO augmentation. 

\begin{table}[t]
	\centering
	\small
    \setlength\tabcolsep{1pt}
	\begin{tabular}{ccccccccc}
	    \toprule
		\multicolumn{1}{c}{} & \multicolumn{2}{c}{O-Duke} & \multicolumn{2}{c}{O-REID} & \multicolumn{2}{c}{P-REID} \\
		\multicolumn{1}{l}{Method} & R@1 & mAP & R@1 & mAP & R@1 & mAP \\
		\hline
		\multicolumn{1}{l}{PCB \cite{sun2018beyond}} & 42.6 & 33.7 & 41.3 & 38.9 & 66.3 & 63.8 \\
		\multicolumn{1}{l}{RE \cite{zhong2020random}} & 40.5 & 30.0 & - & - & 54.3 & 54.4 \\
		\multicolumn{1}{l}{FD-GAN \cite{ge2018fd}} & 40.8 & - & - & - & - & - \\
		\multicolumn{1}{l}{DSR \cite{he2018deep}} & 40.8 & 30.4 & 72.8 & 62.8 & 73.7 & 68.07 \\
		\multicolumn{1}{l}{SFR \cite{he2018recognizing}} & 42.3 & 32 & - & - & 56.9 & - \\
		\multicolumn{1}{l}{FRR \cite{he2019foreground}} & - & - & 78.3 & 68.0 & 81.0 & 76.6 \\
		\multicolumn{1}{l}{PVPM \cite{gao2020pose}} & 47 & 37.7 & 70.4 & 61.2 & - & - \\
		\multicolumn{1}{l}{PGFA \cite{miao2019pose}} & 51.4 & 37.3 & - & - & 69.0 & 61.5 \\
		\multicolumn{1}{l}{HOReID \cite{wang2020high}} & 55.1 & 43.8 & 80.3 & 70.2 & 85.3 & - \\
		\multicolumn{1}{l}{OAMN \cite{chen2021occlude}} & 62.6 & 46.1 & - & - & 86.0 & - \\
		\multicolumn{1}{l}{PAT \cite{li2021diverse}} & 64.5 & 53.6 & 81.6 & 72.1 & \textbf{88.0} & - \\
		\hline
		\multicolumn{1}{l}{ViT Baseline \cite{he2021transreid}} & 60.5 & 53.1 & 81.2 & 76.7 & 73.3 & 74.0 \\
		\multicolumn{1}{l}{TransReID \cite{he2021transreid}} & 64.2 & 55.7 & 70.2 & 67.3 & 71.3 & 68.6 \\
		\hline
		\multicolumn{1}{l}{FED (Ours)} & \textbf{68.1} & \textbf{56.4} & \textbf{86.3} & \textbf{79.3} & 83.1 & \textbf{80.5}\\
		\multicolumn{1}{l}{FED* (Ours)} & 67.9 & 56.3 & \textbf{87.0} & \textbf{79.4} & 84.6 & \textbf{82.3}\\
		\bottomrule
	\end{tabular}
	\caption{Performance comparison with state-of-the-art methods on Occlude-DukeMTMC, Occluded-ReID and Partial-REID datasets. * indicates combining $OS_1$ and $OS_2$ for NPO augmentation. 
	}
	\label{occluded}
\end{table}

\begin{table}[t]
	\centering
	\small
	\begin{tabular}{ccccc}
	    \toprule
		\multicolumn{1}{c}{} & \multicolumn{2}{c}{Market-1501} & \multicolumn{2}{c}{DukeMTMC-reID} \\
		\multicolumn{1}{c}{Model} & Rank-1 & mAP & Rank-1 &  mAP \\
		\hline
		\multicolumn{1}{l}{PT \cite{liu2018poset}} & 87.7 & 68.9 & 78.5 & 56.9 \\
		\multicolumn{1}{l}{PGFA \cite{miao2019pose}} & 91.2 & 76.8 & 82.6 & 65.5 \\
		\multicolumn{1}{l}{PCB \cite{sun2018beyond}} & 92.3 & 77.4 & 81.8 & 66.1 \\
		\multicolumn{1}{l}{OAMN \cite{chen2021occlude}} & 92.3 & 79.8 & 86.3 & 72.6 \\
		\multicolumn{1}{l}{BoT \cite{luo2019bag}} & 94.1 & 85.7 & 86.4 & 76.4 \\

		\multicolumn{1}{l}{HOReID \cite{wang2020high}} & 94.2 & 84.9 & 86.9 & 75.6 \\
		\multicolumn{1}{l}{PAT \cite{li2021diverse}} & \textbf{95.4} & 88.0 & 88.8 & 78.2 \\
		\hline
		\multicolumn{1}{l}{ViT Baseline \cite{he2021transreid}} & 94.7 & 86.8 & 88.8 & 79.3 \\
		\multicolumn{1}{l}{TransReID \cite{he2021transreid}} & 95.0 & \textbf{88.2} & \textbf{89.6} & \textbf{80.6}\\
		\hline
		\multicolumn{1}{l}{FED (Ours)} & 95.0 & 86.3 & 89.4 & 78.0\\
		\bottomrule
	\end{tabular}
	\caption{Performance comparison with state-of-the-art methods on Market-1501 and DukeMTMC-reID datasets.
	}
	\label{holistic}
\end{table}

\subsection{Comparison with State-of-the-art Methods}
\textbf{Comparisons on Occluded Datasets.}
The results on Occluded-DukeMTMC (O-Duke), Occluded-REID (O-REID), and Partial-REID (P-REID) are shown in Table \ref{occluded}. Since O-REID and P-REID don't have  corresponding training set, we simply adopt the model trained on Market-1501 for testing.
PAT \cite{li2021diverse} makes great improvement on accuracy. They adopt ResNet50 \cite{he2016deep} as the backbone and conduct diverse part discovery through the transformer encoder-decoder structure. The prototypes in the network are like specific feature detectors, which are important to improve the network performance on occluded data. 
TransReID \cite{he2021transreid} is the first pure transform-based architecture for person ReID. For a fair comparison, we present the results of TransReID that adopts the Vision Transformer \cite{dosovitskiy2020image} without the sliding window setting as the backbone and images resized to $256 \times 128$. Since He \emph{et al.} \cite{he2021transreid} do not provide performance on O-REID and P-REID datasets, we retrain their official code on Market-1501 dataset and test on the two occluded datasets. The ViT Baseline performs better than TransReID on O-REID and P-REID datasets, this is because TransReID employs many dataset-specific tokens, which reduces the model's cross-domain generalization and increases the overfitting risk.

When comparing our FED (augmented by $OS_1$) with state-of-the-art methods, we achieve the highest Rank-1 and mAP on both O-Duke and O-REID datasets. Especially on O-REID dataset, we achieve 86.3\%/79.3\% on Rank-1/mAP, surpassing others by at least 4.7\%/2.6\%. On O-Duke, we achieve 68.1\%/56.4\% on Rank-1/mAP, surpassing others by at least 3.6\%/0.7\%. 
On the P-REID dataset, we achieve the highest mAP accuracy, reaching 80.5\% and surpassing other methods by 3.9\%. We fail to achieve the highest Rank-1 accuracy on this dataset due to the low generalization of ViT backbone trained on a small dataset. 
Meanwhile, to further demonstrate the flexibility and scalability of the FED, we add more diversified patches (combining $OS_1$ and $OS_2$) for NPO augmentation. As we can see from the table, FED* improves Rank-1/mAP on O-REID and P-REID by at least 0.7\% by simply improving the diversity of the occlusion set. In conclusion, we achieve great performance on the occluded ReID datasets.

\textbf{Comparisons on Holistic Datasets.}
We also experiment on holistic person ReID datasets, including Market-1501 and DukeMTMC-reID. While training on the DukeMTMC-reID dataset, \emph{MSE Loss} is not calculated. It is because huge amounts NPO exist in the training set and we are unable to get precise occlusion masks. 
The results are shown in Table.\ref{holistic}. We achieve comparable performance compared with other state-of-the-art methods.
The same as \emph{Section 4.3.1}, the TransReID is without the sliding window setting and with $256 \times 128$ image size. It is clear that TransReID gets better performance than our method on the holistic datasets. This is because TransReID is specifically designed for holistic ReID and encodes camera information during the training process. Besides, our proposed three components, which aim at tackling the occlusion issues, are not fully functional on holistic ReID datasets. However, we also achieve 84.9\% Rank-1 accuracy on DukeMTMC-reID, surpassing other CNN-based methods and close to TransReID. 

\begin{figure}
  \centering
   \includegraphics[width=0.85\linewidth]{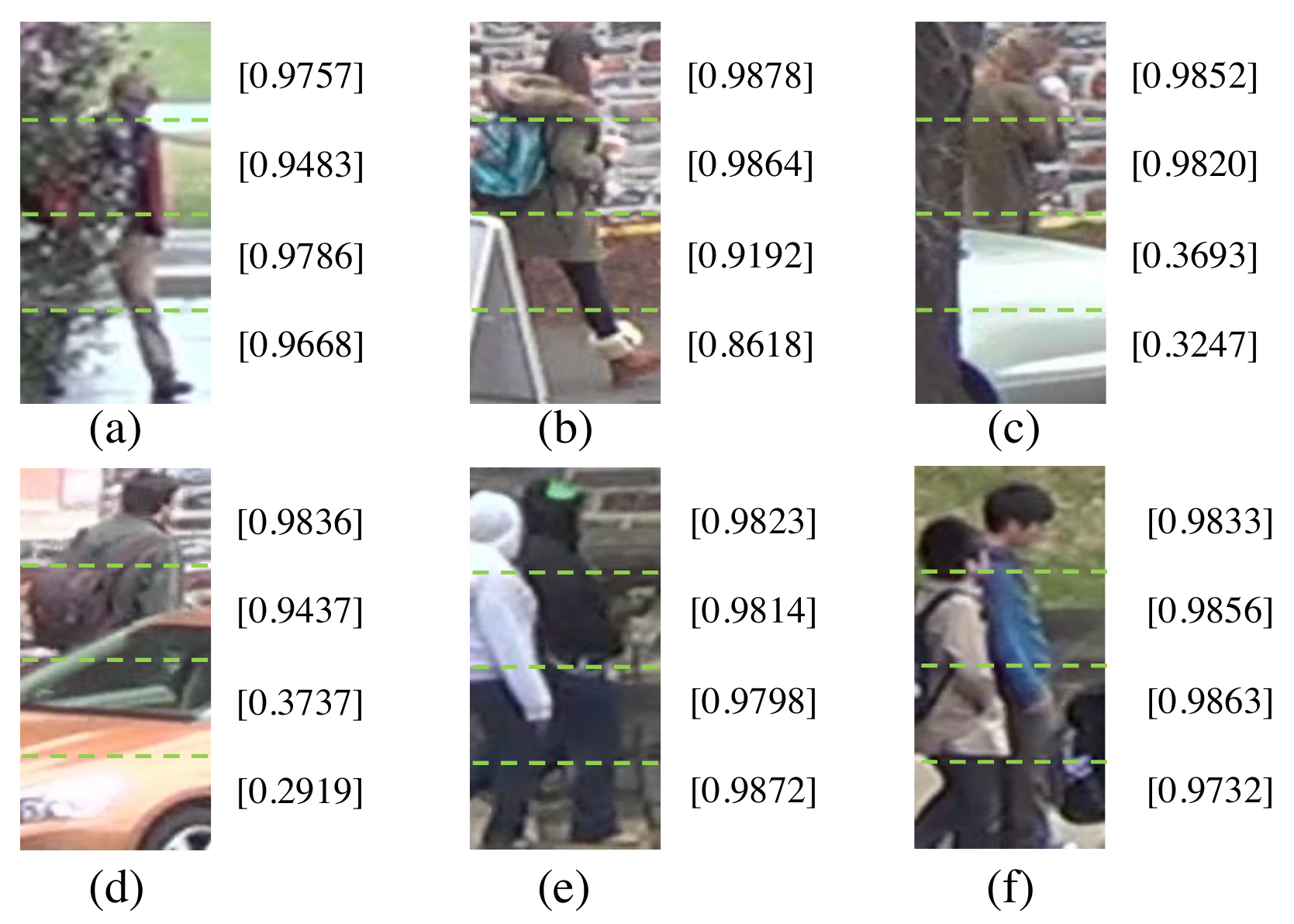}
   \caption{Occlusion scores of OEM on horizontal occluded, vertical occluded and multi-pedestrian images. The OEM has the capacity to identify crucial NPO and fails on NTP.}
   \label{Fig4}
\end{figure}

\begin{figure*}
  \centering
   \includegraphics[width=0.85\linewidth,height=0.28\textwidth]{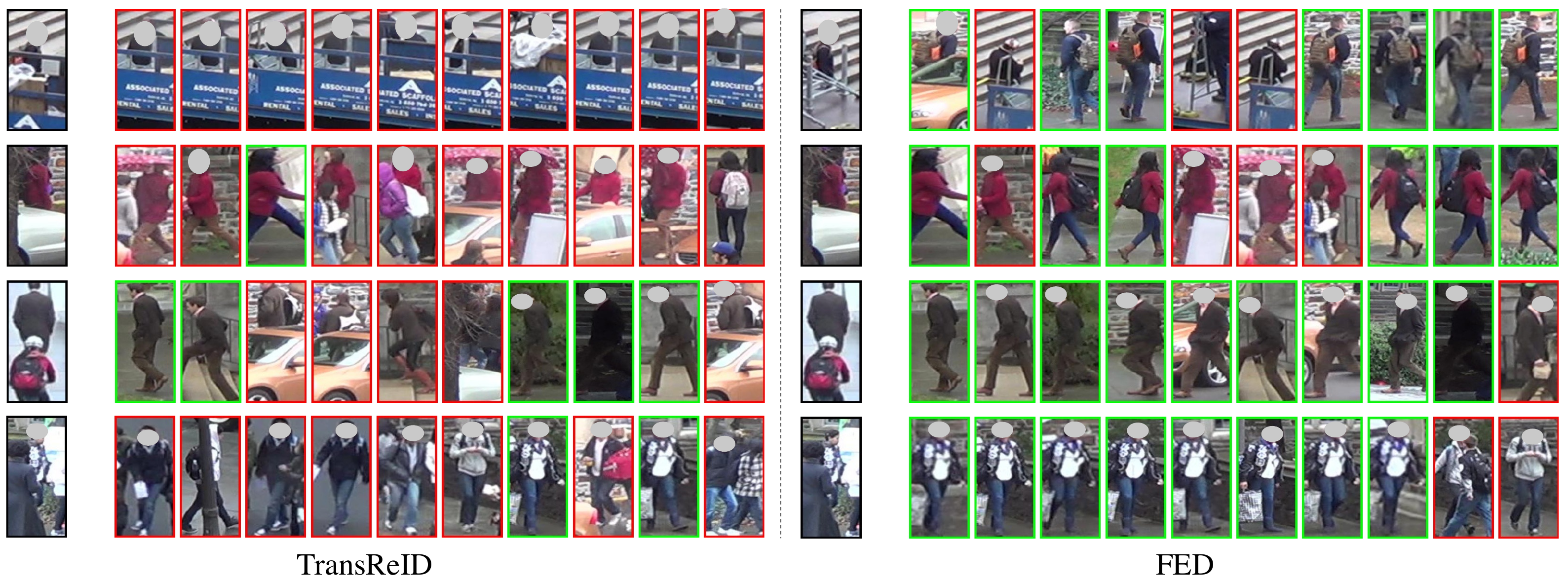}
   \caption{Retrieval results of TransReID and our proposed FED on Occluded-DukeMTMC dataset. The top 2 rows show images with NPO and the bottom 2 rows show images with NTP.}
   \label{Fig6}
\end{figure*}

\begin{table}[t]
	\centering
	\small
	\begin{tabular}{ccccccc}
	    \toprule
		\multicolumn{1}{c}{} & \multicolumn{5}{c}{Occluded-DukeMTMC}  \\
		\hline
		\multicolumn{1}{l}{Index} & RE & NPO Aug & OEM & FDM &  R@1 & mAP \\
		\hline
		\multicolumn{1}{c}{0} & \XSolidBrush & \XSolidBrush  & \XSolidBrush & \XSolidBrush  & 59.1 & 49.1 \\
		\multicolumn{1}{c}{1} & \Checkmark & \XSolidBrush  & \XSolidBrush & \XSolidBrush  & 60.3 & 53.1 \\
		\multicolumn{1}{c}{2} & \XSolidBrush & \Checkmark & \XSolidBrush & \XSolidBrush & 65.4 & 53.5  \\
		\multicolumn{1}{c}{3} & \XSolidBrush & \Checkmark & \Checkmark & \XSolidBrush & 66.5 & 55.4\\
		\multicolumn{1}{c}{4} & \XSolidBrush & \Checkmark & \XSolidBrush & \Checkmark & 67.1 & 55.9  \\
		\multicolumn{1}{c}{5} & \XSolidBrush & \Checkmark & \Checkmark & \Checkmark & 68.1 & 56.4 \\
		\bottomrule
	\end{tabular}
	\caption{Performance analysis of each component in FED. 
	}
	\label{ablation}
\end{table}

\begin{table*}[t]
	\centering
	\small
	\begin{tabular}{cccccccccccc}
	    \toprule
		\multicolumn{1}{c}{} & \multicolumn{3}{c}{Occlude-DukeMTMC} & \multicolumn{3}{c}{DukeMTMC-reID} & \multicolumn{3}{c}{Market-1501} \\
		\multicolumn{1}{l}{Model} & Rank-1 & Rank-5 & mAP & Rank-1 & Rank-5 & mAP & Rank-1 & Rank-5 & mAP \\
		\hline
		\multicolumn{1}{l}{K=2} & 67.4 & 78.4 & 55.8 & \textbf{89.4} & \textbf{94.7} & 77.6 & \textbf{95.0} & \textbf{98.6} & 85.7\\
		\multicolumn{1}{l}{K=4} & 67.7 & \textbf{79.9} & 56.2 & 89.2 & 94.3 & \textbf{78.0} & 94.8 & 98.5 & \textbf{86.3}\\
		\multicolumn{1}{l}{K=6} & 67.3 & 79.8 & 56.2 & 88.9 & 94.2 &77.3 & 94.8 & 98.4 & 86.0\\
		\multicolumn{1}{l}{K=8} & \textbf{68.1} & 79.3 & \textbf{56.4} & 89.0 & 94.3 & 77.1 & 94.8 & 98.4 & 85.9\\
		\bottomrule
	\end{tabular}
	\caption{Analysis of the $K$ in memory searching on both occluded and holistic datasets. CMC curve and mAP are presented for evaluation.
	}
	\label{memorysearch}
\end{table*}

\subsection{Ablation Studies}
\textbf{Analysis of Each Component.} 
In Table.\ref{ablation}, we present the ablation studies of random erasing (RE), NPO augmentation strategy (NPO Aug), occlusion erasing module (OEM), and feature diffusion module (FDM). The indexes from 0 to 5 represent \emph{baseline}, \emph{baseline + RE}, \emph{baseline + NPO Aug}, \emph{baseline + NPO Aug + OEM}, \emph{baseline + NPO Aug + FDM} and \emph{FED}, respectively. All the models adopt ViT as the feature extractor. $Model_0$ and $Model_1$ are both optimized by \emph{ID Loss} and \emph{Triplet Loss} \cite{hermans2017defense}. By comparing $Model_{0\sim2}$, we can see that RE \cite{zhong2020random} is effective in improving discrimination of representations, however the improvement is not comparable with our NPO Aug (4.9\% higher on Rank 1). We can conclude that the augmented images through NPO Aug are realistic and valuable. By comparing $Model_2$ with $Model_3$, the proposed OEM can further improve the representations and improve mAP by 1.9\% by removing the potential NPO information. By comparing $Model_2$ with $Model_4$, FDM helps the model with 1.7\% and 2.4\% improvements on Rank-1 and mAP. It means that optimizing the network with diffused features can greatly improve the model's perception ability towards TP. Finally, FED achieves the highest accuracy, demonstrating that each component can work individually and cooperatively.


\textbf{Analysis of the $K$ in Memory Searching.}
Here, we analyze the searching number $K$ in the memory searching operation. In Table.\ref{memorysearch}, we set $K$ as 2, 4, 6, and 8 and conduct experiments on both holistic and occluded datasets. As we can see, the performance on holistic ReID datasets appears stably on the various $K$, with a float within 0.5\%. For the Market-1501, there are few NPO and NTP, failing to highlight the effectiveness of FDM. For the DukeMTMC-reID, huge amounts of training data are with NPO and NTP, and loss constraints can enable the network with high accuracy. 
As for the Occluded-DukeMTMC, since all the training data are holistic pedestrians, the introduction of FDM can greatly simulate the multi-pedestrian conditions in the testing set. With increasing $K$, FDM can better maintain the characteristics of TP and introduce realistic noise.

\subsection{Qualitative Analysis}
In this section, we present qualitative experimental results and demonstrate the superiority of our proposed FED.

In Fig.\ref{Fig4}, we present the occlusion scores from OEM for some pedestrian images. Images with NPO and NTP are presented. As can be seen, vertical object occlusions (Fig.\ref{Fig4} a, b) can hardly affect the occlusion scores, since occluding less than half of symmetric pedestrians is not a critical issue for person ReID. For horizontal occlusions (Fig.\ref{Fig4} c, d), our OEM can precisely identify NPO and label them with smaller values. For multi-pedestrian images (Fig.\ref{Fig4} e,f), OEM identifies each stripe as valuable. Taken together, the subsequent FDM is essential for improving the model.

In Fig.\ref{Fig6}, we present the retrieval results of TransReID and our FED. The first two examples are object occluded images. It is obvious that our network has a better recognition ability on NPO and accordingly can retrieve target pedestrians precisely. Another two examples provided are the multi-pedestrian images. Our proposed FED has a stronger perception ability on TP and achieves a much higher retrieval accuracy.

\section{Conclusion}
In this paper, to tackle the NPO and NTP challenges for occluded person ReID, we propose a novel Feature Erasing and Diffusion network (FED). Specifically, guided by the image-level NPO augmentation strategy, the occlusion erasing module (OEM) is trained to eliminate NPO features based on the predicted occlusion scores. Subsequently, the feature diffusion module (FDM) performs feature diffusion between NPO-feature-erased pedestrian representations and memorized features, synthesizing NTP characteristics in the feature space. Jointly optimizing OEM and FDM in our proposed FED network significantly improves the model's perception ability on TP, which is demonstrated through comprehensive experiments and comparisons with state-of-the-art algorithms on various person ReID benchmarks.


\newpage

{\small
\bibliographystyle{ieee_fullname}

\begin{thebibliography}{10}
\providecommand{\url}[1]{#1}
\csname url@samestyle\endcsname
\providecommand{\newblock}{\relax}
\providecommand{\bibinfo}[2]{#2}
\providecommand{\BIBentrySTDinterwordspacing}{\spaceskip=0pt\relax}
\providecommand{\BIBentryALTinterwordstretchfactor}{4}
\providecommand{\BIBentryALTinterwordspacing}{\spaceskip=\fontdimen2\font plus
\BIBentryALTinterwordstretchfactor\fontdimen3\font minus
  \fontdimen4\font\relax}
\providecommand{\BIBforeignlanguage}[2]{{%
\expandafter\ifx\csname l@#1\endcsname\relax
\typeout{** WARNING: IEEEtran.bst: No hyphenation pattern has been}%
\typeout{** loaded for the language `#1'. Using the pattern for}%
\typeout{** the default language instead.}%
\else
\language=\csname l@#1\endcsname
\fi
#2}}
\providecommand{\BIBdecl}{\relax}
\BIBdecl

\bibitem{chen2021occlude}
P.~Chen, W.~Liu, P.~Dai, J.~Liu, Q.~Ye, M.~Xu, Q.~Chen, and R.~Ji, ``Occlude
  them all: Occlusion-aware attention network for occluded person re-id,'' in
  \emph{ICCV}, 2021, pp. 11\,833--11\,842.

\bibitem{zhong2020random}
Z.~Zhong, L.~Zheng, G.~Kang, S.~Li, and Y.~Yang, ``Random erasing data
  augmentation,'' in \emph{AAAI}, vol.~34, no.~07, 2020, pp. 13\,001--13\,008.

\bibitem{he2021transreid}
S.~He, H.~Luo, P.~Wang, F.~Wang, H.~Li, and W.~Jiang, ``Transreid:
  Transformer-based object re-identification,'' \emph{ICCV}, 2021.

\bibitem{ge2020self}
Y.~Ge, F.~Zhu, D.~Chen, R.~Zhao, and H.~Li, ``Self-paced contrastive learning
  with hybrid memory for domain adaptive object re-id,'' \emph{NIPS}, 2020.

\bibitem{sun2019deep}
K.~Sun, B.~Xiao, D.~Liu, and J.~Wang, ``Deep high-resolution representation
  learning for human pose estimation,'' in \emph{CVPR}, 2019, pp. 5693--5703.

\bibitem{zhao2017pyramid}
H.~Zhao, J.~Shi, X.~Qi, X.~Wang, and J.~Jia, ``Pyramid scene parsing network,''
  in \emph{CVPR}, 2017, pp. 2881--2890.

\bibitem{vaswani2017attention}
A.~Vaswani, N.~Shazeer, N.~Parmar, J.~Uszkoreit, L.~Jones, A.~N. Gomez,
  {\L}.~Kaiser, and I.~Polosukhin, ``Attention is all you need,'' in
  \emph{NIPS}, 2017, pp. 5998--6008.

\bibitem{xiong2020layer}
R.~Xiong, Y.~Yang, D.~He, K.~Zheng, S.~Zheng, C.~Xing, H.~Zhang, Y.~Lan,
  L.~Wang, and T.~Liu, ``On layer normalization in the transformer
  architecture,'' in \emph{ICML}.\hskip 1em plus 0.5em minus 0.4em\relax PMLR,
  2020, pp. 10\,524--10\,533.

\bibitem{miao2019pose}
J.~Miao, Y.~Wu, P.~Liu, Y.~Ding, and Y.~Yang, ``Pose-guided feature alignment
  for occluded person re-identification,'' in \emph{ICCV}, 2019, pp. 542--551.

\bibitem{zheng2015partial}
W.-S. Zheng, X.~Li, T.~Xiang, S.~Liao, J.~Lai, and S.~Gong, ``Partial person
  re-identification,'' in \emph{ICCV}, 2015, pp. 4678--4686.

\bibitem{he2018deep}
L.~He, J.~Liang, H.~Li, and Z.~Sun, ``Deep spatial feature reconstruction for
  partial person re-identification: Alignment-free approach,'' in \emph{CVPR},
  2018, pp. 7073--7082.

\bibitem{zheng2015scalable}
L.~Zheng, L.~Shen, L.~Tian, S.~Wang, J.~Wang, and Q.~Tian, ``Scalable person
  re-identification: A benchmark,'' in \emph{ICCV}, 2015, pp. 1116--1124.

\bibitem{zheng2017unlabeled}
Z.~Zheng, L.~Zheng, and Y.~Yang, ``Unlabeled samples generated by gan improve
  the person re-identification baseline in vitro,'' in \emph{ICCV}, 2017, pp.
  3754--3762.

\bibitem{zhuo2018occluded}
J.~Zhuo, Z.~Chen, J.~Lai, and G.~Wang, ``Occluded person re-identification,''
  in \emph{ICME}.\hskip 1em plus 0.5em minus 0.4em\relax IEEE, 2018, pp. 1--6.

\bibitem{hermans2017defense}
A.~Hermans, L.~Beyer, and B.~Leibe, ``In defense of the triplet loss for person
  re-identification,'' \emph{arXiv preprint arXiv:1703.07737}, 2017.

\bibitem{luo2019bag}
H.~Luo, Y.~Gu, X.~Liao, S.~Lai, and W.~Jiang, ``Bag of tricks and a strong
  baseline for deep person re-identification,'' in \emph{CVPRW}, 2019, pp.
  0--0.

\bibitem{he2019foreground}
L.~He, Y.~Wang, W.~Liu, H.~Zhao, Z.~Sun, and J.~Feng, ``Foreground-aware
  pyramid reconstruction for alignment-free occluded person
  re-identification,'' in \emph{ICCV}, 2019, pp. 8450--8459.

\bibitem{he2020guided}
L.~He and W.~Liu, ``Guided saliency feature learning for person
  re-identification in crowded scenes,'' in \emph{ECCV}.\hskip 1em plus 0.5em
  minus 0.4em\relax Springer, 2020, pp. 357--373.

\bibitem{yang2014salient}
Y.~Yang, J.~Yang, J.~Yan, S.~Liao, D.~Yi, and S.~Z. Li, ``Salient color names
  for person re-identification,'' in \emph{ECCV}.\hskip 1em plus 0.5em minus
  0.4em\relax Springer, 2014, pp. 536--551.

\bibitem{ma2014covariance}
B.~Ma, Y.~Su, and F.~Jurie, ``Covariance descriptor based on bio-inspired
  features for person re-identification and face verification,'' \emph{Image
  and Vision Computing}, vol.~32, no. 6-7, pp. 379--390, 2014.

\bibitem{chen2017beyond}
W.~Chen, X.~Chen, J.~Zhang, and K.~Huang, ``Beyond triplet loss: a deep
  quadruplet network for person re-identification,'' in \emph{CVPR}, 2017, pp.
  403--412.

\bibitem{zhong2017re}
Z.~Zhong, L.~Zheng, D.~Cao, and S.~Li, ``Re-ranking person re-identification
  with k-reciprocal encoding,'' in \emph{CVPR}, 2017, pp. 1318--1327.

\bibitem{wang2020robust}
Z.~Wang, L.~He, X.~Gao, and J.~Shen, ``Robust person re-identification through
  contextual mutual boosting,'' \emph{arXiv preprint arXiv:2009.07491}, 2020.

\bibitem{sun2018beyond}
Y.~Sun, L.~Zheng, Y.~Yang, Q.~Tian, and S.~Wang, ``Beyond part models: Person
  retrieval with refined part pooling (and a strong convolutional baseline),''
  in \emph{ECCV}, 2018, pp. 480--496.

\bibitem{xu2018attention}
J.~Xu, R.~Zhao, F.~Zhu, H.~Wang, and W.~Ouyang, ``Attention-aware compositional
  network for person re-identification,'' in \emph{CVPR}, 2018, pp. 2119--2128.

\bibitem{zhang2017alignedreid}
X.~Zhang, H.~Luo, X.~Fan, W.~Xiang, Y.~Sun, Q.~Xiao, W.~Jiang, C.~Zhang, and
  J.~Sun, ``Alignedreid: Surpassing human-level performance in person
  re-identification,'' \emph{arXiv preprint arXiv:1711.08184}, 2017.

\bibitem{li2018harmonious}
W.~Li, X.~Zhu, and S.~Gong, ``Harmonious attention network for person
  re-identification,'' in \emph{CVPR}, 2018, pp. 2285--2294.

\bibitem{tay2019aanet}
C.-P. Tay, S.~Roy, and K.-H. Yap, ``Aanet: Attribute attention network for
  person re-identifications,'' in \emph{CVPR}, 2019, pp. 7134--7143.

\bibitem{yang2019attention}
F.~Yang, K.~Yan, S.~Lu, H.~Jia, X.~Xie, and W.~Gao, ``Attention driven person
  re-identification,'' \emph{Pattern Recognition}, vol.~86, pp. 143--155, 2019.

\bibitem{huang2020human}
H.~Huang, X.~Chen, and K.~Huang, ``Human parsing based alignment with
  multi-task learning for occluded person re-identification,'' in
  \emph{ICME}.\hskip 1em plus 0.5em minus 0.4em\relax IEEE, 2020, pp. 1--6.

\bibitem{ba2016layer}
J.~L. Ba, J.~R. Kiros, and G.~E. Hinton, ``Layer normalization,'' \emph{arXiv
  preprint arXiv:1607.06450}, 2016.

\bibitem{ge2018fd}
Y.~Ge, Z.~Li, H.~Zhao, G.~Yin, S.~Yi, X.~Wang, and H.~Li, ``Fd-gan: Pose-guided
  feature distilling gan for robust person re-identification,'' in \emph{NIPS},
  2018, pp. 1229--1240.

\bibitem{he2018recognizing}
L.~He, Z.~Sun, Y.~Zhu, and Y.~Wang, ``Recognizing partial biometric patterns,''
  \emph{arXiv preprint arXiv:1810.07399}, 2018.

\bibitem{gao2020pose}
S.~Gao, J.~Wang, H.~Lu, and Z.~Liu, ``Pose-guided visible part matching for
  occluded person reid,'' in \emph{CVPR}, 2020, pp. 11\,744--11\,752.

\bibitem{wang2020high}
G.~Wang, S.~Yang, H.~Liu, Z.~Wang, Y.~Yang, S.~Wang, G.~Yu, E.~Zhou, and
  J.~Sun, ``High-order information matters: Learning relation and topology for
  occluded person re-identification,'' in \emph{CVPR}, 2020, pp. 6449--6458.

\bibitem{li2021diverse}
Y.~Li, J.~He, T.~Zhang, X.~Liu, Y.~Zhang, and F.~Wu, ``Diverse part discovery:
  Occluded person re-identification with part-aware transformer,'' in
  \emph{CVPR}, 2021, pp. 2898--2907.

\bibitem{he2016deep}
K.~He, X.~Zhang, S.~Ren, and J.~Sun, ``Deep residual learning for image
  recognition,'' in \emph{CVPR}, 2016, pp. 770--778.

\bibitem{dosovitskiy2020image}
A.~Dosovitskiy, L.~Beyer, A.~Kolesnikov, D.~Weissenborn, X.~Zhai,
  T.~Unterthiner, M.~Dehghani, M.~Minderer, G.~Heigold, S.~Gelly \emph{et~al.},
  ``An image is worth 16x16 words: Transformers for image recognition at
  scale,'' \emph{arXiv preprint arXiv:2010.11929}, 2020.

\bibitem{wang2021robust}
Z.~Wang, L.~He, X.~Tu, J.~Zhao, X.~Gao, S.~Shen, and J.~Feng, ``Robust
  video-based person re-identification by hierarchical mining,'' \emph{TCSVT},
  2021.

\bibitem{liu2018poset}
J.~Liu, B.~Ni, Y.~Yan, P.~Zhou, S.~Cheng, and J.~Hu, ``Pose transferrable
  person re-identification,'' in \emph{CVPR}, 2018, pp. 4099--4108.

\bibitem{yu2021neighbourhood}
S.~Yu, D.~Chen, R.~Zhao, H.~Chen, and Y.~Qiao, ``Neighbourhood-guided feature
  reconstruction for occluded person re-identification,'' \emph{arXiv preprint
  arXiv:2105.07345}, 2021.

\bibitem{wei2018person}
L.~Wei, S.~Zhang, W.~Gao, and Q.~Tian, ``Person transfer gan to bridge domain
  gap for person re-identification,'' in \emph{CVPR}, 2018, pp. 79--88.

\bibitem{wang2019multi}
Z.~Wang, L.~He, X.~Gao, and Y.~Huang, ``Multi-scale spatial-temporal network
  for person re-identification,'' in \emph{ICASSP}.\hskip 1em plus 0.5em minus
  0.4em\relax IEEE, 2019, pp. 2052--2056.

\end{thebibliography}

}

\end{document}